\documentclass[letterpaper, 10 pt, conference]{ieeeconf}  
\usepackage{graphicx}
\usepackage{array}
\usepackage{tabularx}
\usepackage{multirow}
\usepackage{siunitx}
\usepackage{booktabs}
\usepackage{floatrow}
\usepackage{blindtext}
\usepackage{subcaption}
\usepackage{hyperref}
\hypersetup{
  colorlinks   = true, 
  urlcolor     = blue, 
  linkcolor    = blue, 
  citecolor   = red 
}

\IEEEoverridecommandlockouts                              

\usepackage{atbegshi}
\AtBeginDocument{\AtBeginShipoutNext{\AtBeginShipoutDiscard}}

\begin{document}

\title{\LARGE \bf Real-Time Heuristic Framework for Safe Landing of UAVs in Dynamic Scenarios
}

\author{Jaskirat Singh$^{\ast}$$^{1}$, Neel Adwani$^{\ast}$$^{1}$, Harikumar Kandath$^{2}$, and K. Madhava Krishna$^{2}$}

\thanks{$^{*}$ Denote equal contribution}%
\thanks{$^{1}$University of Petroleum and Energy Studies, Dehradun
 {\tt\small {juskirat2000@gmail.com, neeltr.n@gmail.com}}}

\thanks{$^{2}$International Institute of Information Technology, Hyderabad
{\tt\small {harikumar.k@iiit.ac.in, mkrishna@iiit.ac.in}}}

\maketitle
\thispagestyle{empty}
\pagestyle{empty}

\begin{abstract}
The world we live in is full of technology and with each passing day the advancement and usage of UAVs increases efficiently. As a result of the many application scenarios, there are some missions where the UAVs are vulnerable to external disruptions, such as a ground station's loss of connectivity, security missions, safety concerns, and delivery-related missions. Therefore, depending on the scenario, this could affect the operations and result in the safe landing of UAVs. Hence,
this paper presents a heuristic approach towards safe landing of multi-rotor UAVs in the dynamic environments. The aim of this approach is to detect safe potential landing zones (PLZ), and find out the best one to land in. The PLZ is initially, detected by processing an image through the canny edge algorithm, and then the diameter-area estimation is applied for each region with minimal edges. The spots that have a higher area than the vehicle's clearance are labeled as safe PLZ. Onto the second phase of this approach, the velocities of dynamic obstacles that are moving towards the PLZs are calculated and their time to reach the zones are taken into consideration. The ETA of the UAV is calculated and during the descending of UAV, the dynamic obstacle avoidance is executed. The approach tested on the real-world environments have shown better results from existing work.\\

\end{abstract}



\section{INTRODUCTION}

The development of autonomous Unmanned Aerial Vehicles (UAVs), which can self-navigate in a range of situations, has been aided by recent advances in artificial intelligence, control, and remote sensing technology. The utilisation of UAVs has increased significantly, showing applications in surveillance and security systems \cite{6127543, 10.1007/978-3-540-76928-6_1}, delivering of products \cite{6919154}, monitoring forest fires \cite{SUDHAKAR20201}, motion and traffic analysis and various other research purposes \cite{6290694, 8939254}. \\
Given the application scenarios, there are missions where the UAVs are susceptible to external disturbances like loss of communication from ground stations, security missions, safety, and delivery reasons. Hence, in the event of different scenarios, this could impact the operations, thus leading to the safe landing of UAVs. In these situations, UAVs must safely land at unprepared sites to reduce damage to themselves and avoid causing any injury to humans. To ensure the safe landing of the UAV using the existing autonomous landing system, it is typically essential to choose a fixed and safe landing location, and ensure that the region is relatively open, flat, has sufficient area and does not hit any dynamic obstacles and people. However, in most cases, during run-time, UAVs do not have prior information about the region of the potential landing area, and the dynamic environments as a best possible landing spots taking into consideration the priority of where to land in the cases of different aborted missions. The effectiveness of the entire unmanned autonomous operation is significantly impacted by the requirement for people to monitor the landing situation or employ manual remote control to land in dynamic circumstances with any moving objects or people. However, in the context of dynamic surroundings, contemporary UAV systems lack suitable landing response techniques. A considerable amount of research has been dedicated in detecting the landing sites with the objects at statically positioned using various computer vision techniques \cite{cv1, cv2, cv3}. This ensures efficient autonomous detection of UAV landing zones at cheaper cost and with less need for human interaction. Although these methods operate well in some specific scenarios, it might be challenging to reliably and properly identify the landing region in more complicated such as dynamic environments where any objects or people are in continuous different motions.

\subsection{Contributions}
The main contributions of this paper are listed below:
\begin{enumerate}
    \item Considering the robustness of problems that a UAV faces during the autonomous safe landing, we primarily have developed the state estimation mechanism to land safely in the static and dynamic environments where any objects or people are in different motions by tracking their velocities, and estimated time of arrival. As far as we are aware, no research has been done that specifically addresses this issue in dynamic circumstances. Compared to the previous works in \cite{KALJAHI2019319, 9811924} have followed the static detection methods.
    
    \item Consequently, detecting a reliable potential landing zone is essential for safe operations. Hence to estimate the potential landing zone, a set of conditions have to be evaluated when analyzing the different sensor data. Therefore, we introduced the architecture in Figure \ref{fig:architecture} for finding the area of potential landing zone (PLZ), and distance of the UAV to the PLZ. To the best of our knowledge, we didn’t find any work that calculates the resulting area of potential landing zone for UAVs without state-of-art deep learning networks. We have also compared our results with some methods where our method has shown great accuracy and precision for area identification.
    
    \item Furthermore, we evaluated our proposed approach with field tests in real-world dynamic environments.
    
    \item We developed an open source framework for safe landing of UAVs in dynamic environments that can be embedded into any micro-processor for future work.
    
\end{enumerate}

\section{RELATED WORK}

\subsection{Potential Landing Zone Detection}

Despite the increasing importance given to safety, safe landing of drones still remains an open problem, especially in unknown environments \cite{7899875, electronics7050073}. Detecting a safe landing area and approaching it are both challenging tasks, since there are multiple factors involved while detecting the potential landing zones. Thus, its is essential to process different types of data that has been collected from several sensors in order to obtain best possible landing zones. Consequently, the recent research regarding the safe landing, either for emergencies or for other purposes have been divided into two different categories; Sensor based detection system \cite{sb1, sb2, sb3, sb4}, and Vision based detection system \cite{vision1, vision2}. The authors in \cite{9507275} created a density map for each image using a deep neural network, and obtained a binary occupancy map aiming to overestimate the people's location. Density maps are obtained
by means of DNN trained utilizing crowd images containing head annotations which is one of the drawbacks of using the DNN based networks. Also, the research conducted by G. Castellano et al. in \cite{9052702} have proposed a method for identifying the safe landing zones which is based on light-weight scheme of a fully state-of-the-art CNN networks. Some other approaches have aimed at identifying ‘‘safe’’ areas for possible landing. This is the case, where the work by Mukadam et al. in \cite{7860044} have followed a more conventional approach while making the use of SVM based algorithms to detect potential landing zones by extracting the features from colored satellite images.
Several literature review were studied where the work has primarily been focused around neural networks which includes the training on some set of images.

\subsection{Real-Time Potential Landing Zone State Estimation and Navigation in Dynamic Scenarios}

We couldn't find much related work as a part of literature which attempts to solve the complicated task for PLZ state estimation, taking into consideration the decision-making process for safe landing of UAVs but mostly we could find the literature for the work that helps in "avoiding" dynamic obstacles without defining a path. The authors in \cite{Sanchez-Lopez2019} proposed a probabilistic graph approach, further developed a cost function as to obtain a raw optimal  collision-free path which was tested in a simulation environment. The research conducted by C. Lyujie et al. in \cite{Chen2020RobustAL} offered a way combining inexpensive sensors like binoculars and LiDAR to achieve autonomous landing in hostile settings. By utilising the LIDAR's FOV coverage properties, a dynamic temporal depth completion algorithm was developed.\\
The current effort is concentrated on creating an obstacle avoidance system that can lessen or eliminate the chance of a UAV colliding with things or people that get in the way of the planned path or trajectory mission and fully deviates from it, losing track of trajectories in the process. There have been many different approaches that have been
proposed in recent years on this subject where major approaches have been based on geometric relations \cite{Guo2021, Sasongko2017}, fuzzy logic \cite{Singh_2019, IJOST24889}, and neural networks \cite{8928110, DAI2020346},
among others.  The authors in \cite{drones6010016} proposes an algorithm for navigating through way-points while avoiding obstacles through simplified form of geometry generated from a point cloud of the different scenarios. Similarly authors in \cite{9019059} represents the time-obstacle dynamic map (TODM) to avoid dynamic obstacles.\\

In contrast to the aforementioned techniques mentioned in the literature, the approach presented in this paper detects the potential landing zones and, formulates the path for the UAV, by tracking the velocities, and estimated time of each object to reach the targeted location.\\
Finally, we evaluate our framework in the real-world environments.

\section{METHODOLOGY}

\subsection{System Overview}

\begin{figure*}
    \centering
    \includegraphics[width=\textwidth]{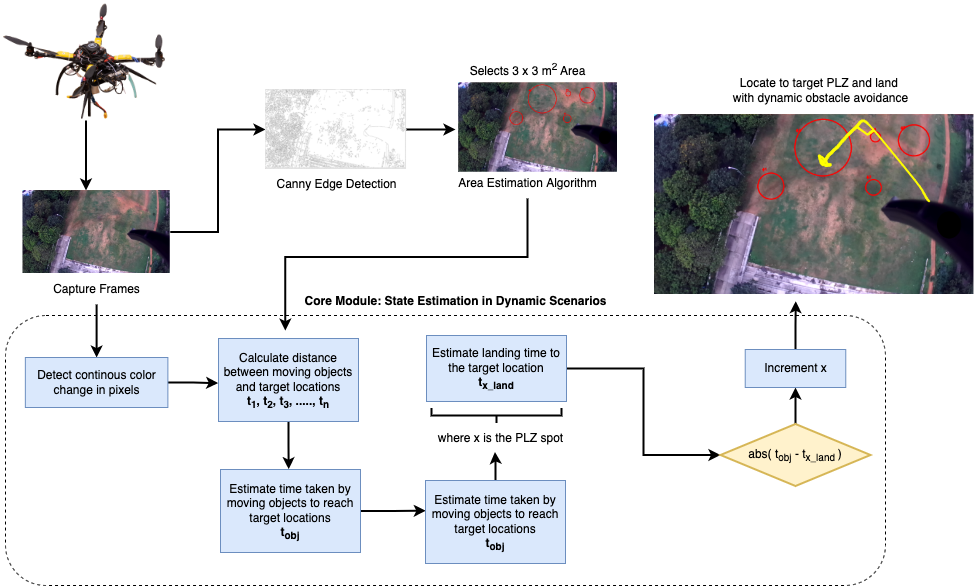}
    \caption{Architecture of the \textit{Safe Landing of UAVs}. UAV capture frames which are input for the \textit{Area Estimation} and \textit{Core Module} that initiates the auto landing.}
    \label{fig:architecture}
\end{figure*}

In this section we present the overview of the proposed system. Hence the overview system of the autonomous safe landing of UAV in dynamic scenarios is presented in the Figure \ref{fig:architecture}. During on-flight the images are captured at 30fps with the help of ZED\footnote{ZED camera specifications and details can be found at the official Stereolabs website: \href{https://www.stereolabs.com/}{https://www.stereolabs.com/}} depth stereo camera where each frame is given as an input to the Canny Edge detection algorithm \cite{4767851}. Diameter-Area estimation algorithm as described in Section IIA is applied on the canny edge binary output image which forms the different possible circles on finding the empty spot, having no edges and the circle having the minimum 3 meters sq. area is  referred as Potential landing Zone (PLZ). At the same time, within the input image, moving objects are detected using the color change in pixels. For every moving object, its distance to PLZ, and velocity is calculated, and stored. Time taken by the object ($T_x$), and quadrotor to reach to nearest PLZ, is taken into consideration.\\
The estimated time of arrival (ETA) of the drone to reach that PLZ spot is calculated by translating the pixels into distance in X-axis, and Y-axis, and considering the altitude in Z-axis. We put on the deciding factor, considering if ($T_x$ - ETA) is greater than 20 seconds, that denotes the specific PLZ is cleared for landing. The auto-land command is initiated as a part of autopilot mode, the drone moves to the position in X-axis, and Y-axis. At the time of descent in Z-Axis, the real-time obstacle avoidance algorithm, divides the present frame into 4 quadrants that makes an occupancy grid map \cite{6907758} where every quadrant contains the depth matrix of every pixel. The average depth is taken from the matrix from each quadrant and is stored into an array. The quadrant with the highest depth is considered as the emptiest quadrant and the UAV would descend towards that quadrant. The on-board Time-of-Flight (ToF) based range measuring sensor\footnote{The ToF sensor can be found at: \href{https://www.terabee.com/shop/lidar-tof-range-finders/teraranger-evo-60m/}{https://www.terabee.com/shop/lidar-tof-range-finders/teraranger-evo-60m/}}, verifies the height of the drone, along with the average depth from the depth camera. If the distance doesn't match, it means that there's an obstacle beyond it, which is the case when drone will move to a different location, re-adjust itself, and descend further. Finally, the UAV lands at the desired PLZ location.

\subsection{Potential Landing Zone Detection} 
As shown in Figure \ref{fig:architecture}, we divided our method into two parts in order to locate the PLZ. In, \textit{Stage I}, just after the drone take-off the image are captured with the help of ZED stereo-depth camera integrated into a drone at 30 FPS, and as a part of continuous evaluation, the frames gets stored in the secure digital (SD) card, embedded into a Jetson Nano. The colour image is first converted to a grayscale image, and the grayscale image is then subjected to the clever edge detection method utilising 50 and 150 as the lower and upper thresholds, respectively. The image's gradient is used by canny edge detection to locate the edges. Utilizing a Gaussian filter's derivative, the gradient of the image is determined. With edge pixels designated by 1 and non-edge pixels denoted by 0, the binary picture produced by the canny edge is its output..\\

In \textit{Stage II}, we apply the diameter-area estimation algorithm. Clustering is done by applying euclidean distance between the contours. Contours having the distance less than 30 pixels is clustered into one set and polygon is formed. The shortest distance between two different polygons is calculated using the Equation \ref{distance-equation} which gives the distance between the two edges ($D_{image}$). where $P_{12}$, and $P_{22}$ are the coordinates of one edge end, and $P_{11}$, and $P_{21}$ are the coordinates of the other end of the edge.

\begin{equation}\label{distance-equation}
    D_{image}\,(m) = \sqrt[2]{(P_{12} - P_{11})^2 + (P_{22} - P_{21})^2}
\end{equation}

Hence, the distance between the two objects in real world ($D_{object}$), can be found using the Equation \ref{distance-realworld-equation} where \textit{A} is the height of the drone from the surface, and \textit{f} is the focal length of camera.

\begin{equation}\label{distance-realworld-equation}
    D_{object}\,(m) = \frac{ D_{image}\times A}{f}
\end{equation}

With the help of Equation \ref{distance-equation}, \ref{distance-realworld-equation}, we can calculate the non-edge area using the Equation \ref{area-equation}

\begin{equation} \label{area-equation}
    A_{PLZ}\,(m^2) = D_{object}.D_{object}
\end{equation}

\begin{figure}
    \includegraphics[scale=0.4]{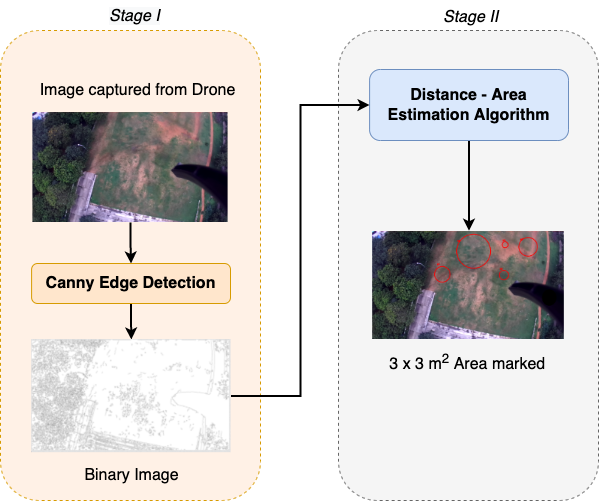}
    \caption{Architecture of potential landing zone detection.}
    \label{fig:dynamicscene}
\end{figure}

\subsection{Real-Time Potential Landing Zone State Estimation and Navigation in Dynamic Scenarios}

The overview of real-time PLZ state estimation, and navigation in dynamic scenario has been described in the Figure \ref{fig:dynamicscene}. The approach for dynamic scenario gets divided into two stages. \textit{Stage I} is a \textbf{Descision-Making} process where the images are captured from the drone, each frame is evaluated for considering the dynamic motions of objects. Understanding the more precise scenario, moving objects ($O_{1}$, $O_{2}$, $O_{3}$, $O_{4}$,....., $O_{n}$) are detected using the color change in pixels. PLZ are represented by ($t_{1}$, $t_{2}$, $t_{3}$, $t_{4}$,....., $t_{n}$). For every PLZ ($t_{x}$), distance from each object ($S_{1,X}$, $S_{2,X}$, $S_{3,X}$, $S_{4,X}$,....., $S_{n,X}$), located in the same axis is calculated using the Equation \ref{distance-equation}. The velocity of every object ($V_{1,x}$, $V_{2,x}$, $V_{3,x}$, $V_{4,x}$,....., $V_{n,x}$) is calculated through the amount of pixels that are shifting in every second. The time taken ($T_{x}$) by the moving object ($O_{m}$) to reach its nearest PLZ ($t_{s}$) is calculated using the Equation \ref{time-equation} where, ($S_{obj,x}$) is the distance between object and target location, and ($V_{obj,x}$) is the velocity of an object to reach target location.

\begin{equation}\label{time-equation}
    T_{x}\,(seconds) = \frac{S_{obj,X}}{V_{obj,x}}
\end{equation}

The ETA of the drone ($T_{d}$) to the target PLZ is calculated by translating the pixels into distances using the Equation \ref{distance-equation} in X-axis, and Y-axis, and with the help of ToF sensor altitude is calculated in Z-axis. If the absolute value of ($T_{x}$ - $T_{d}$) is greater than 20 seconds, that indicates that the specific PLZ ($t_{x}$) is cleared for landing, else the quadrotor slows down the speed in-order to get the clear landing, while waiting the moving objects to pass through the target location.

\begin{figure}
    \includegraphics[scale=0.4]{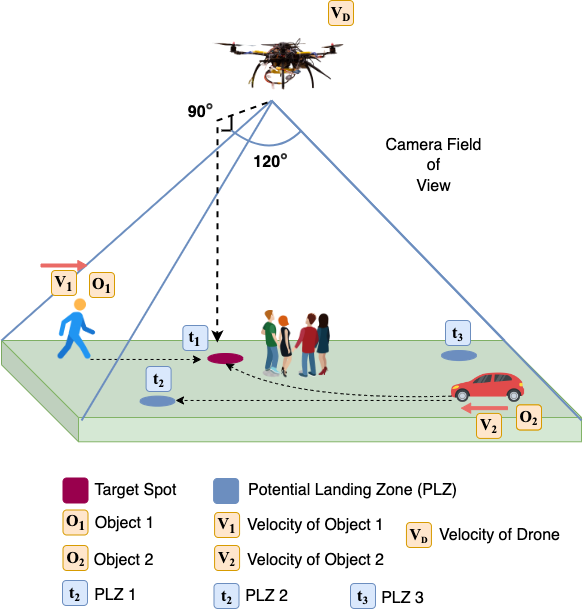}
    \caption{Architecture of real-time potential landing zone in dynamic scenario where objects are in different dynamic motions. As seen person and car are in motion while defining their actual trajectory, passing through target PLZ.}
    \label{fig:dynamicscene}
\end{figure}
\begin{figure}
    \includegraphics[scale=0.4]{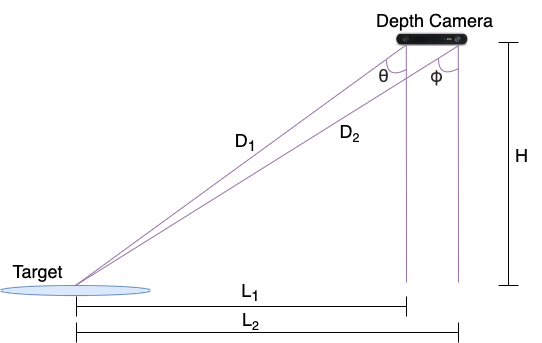}
    \caption{Triangulation for measuring the depth of different quadrants in a frame}
    \label{fig:dynamic-obstaclescene}
\end{figure}

The \textit{Stage II} is \textbf{Navigation}. After the drone has taken decision to land at specific PLZ, it navigates in X-axis and Y-axis such that the target PLZ accounts in the center position of the frame. When the landing of the drone is initiated, i.e. at the time of descent, the real-time obstacle avoidance algorithm acts, dividing the each frame that is taken from the depth camera embedded into a drone into a four quadrants and make an occupancy map. The occupancy grid map shows a map of the area as a uniformly spaced field of binary random variables, where each variable indicates whether a specific obstruction exists within that location around that area. For these random variables, occupancy grid techniques create rough posterior estimates. Every quadrant of the frame contains the depth matrix of every pixel. The depth of the pixel is found using the triangulation rule which has been shown in Figure \ref{fig:dynamic-obstaclescene}. When the distance of the drone to the target is large enough as compared to two different camera's view, then the angle ($\theta$ = $\phi$), the distance ($D_{1}$ = $D_{2}$), distance between the depth camera and target ($L_{1}$ = $L_{2}$). Hence using the Equation \ref{pythagoras-equation} we can find the depth of the specific pixel in a quadrant where, \emph{H} is altitude of the drone from the plane surface, \emph{D} represents the depth of pixel.

\begin{equation}\label{pythagoras-equation}
    D = \frac{H}{\cos \theta}
\end{equation}
The average depth taken from the matrix from each quadrant is stored into an array. This generally denotes that the quadrant having the highest depth is considered as the emptiest depth, indicating UAV to descend towards that specific quadrant. The on-board ToF sensor is used to verify the altitude of the drone, along with he depth from depth camera.  If the distance doesn't match, it means that there's an obstacle beyond it, which is the case when drone will move to a different location and re-adjust itself and descend further.

\section{EXPERIMENTS AND RESULTS}

This section presents the extensive experiments conducted in real-world environments and results were evaluated for different utility scenarios. The proposed framework is tested on a S-500 model quadrotor for safe landing of UAV. The quadrotor is controlled by an onboard Jetson Nano, and Pixhawk PX4 flight controller, with the depth images, and altitude range measure obtained from Depth camera, and ToF sensor respectively.

\subsection{Potential Landing Zone Detection}

\begin{figure*}
  \centering
  
  \begin{subfigure}[b]{.4\textwidth}
  \includegraphics[width=\textwidth]{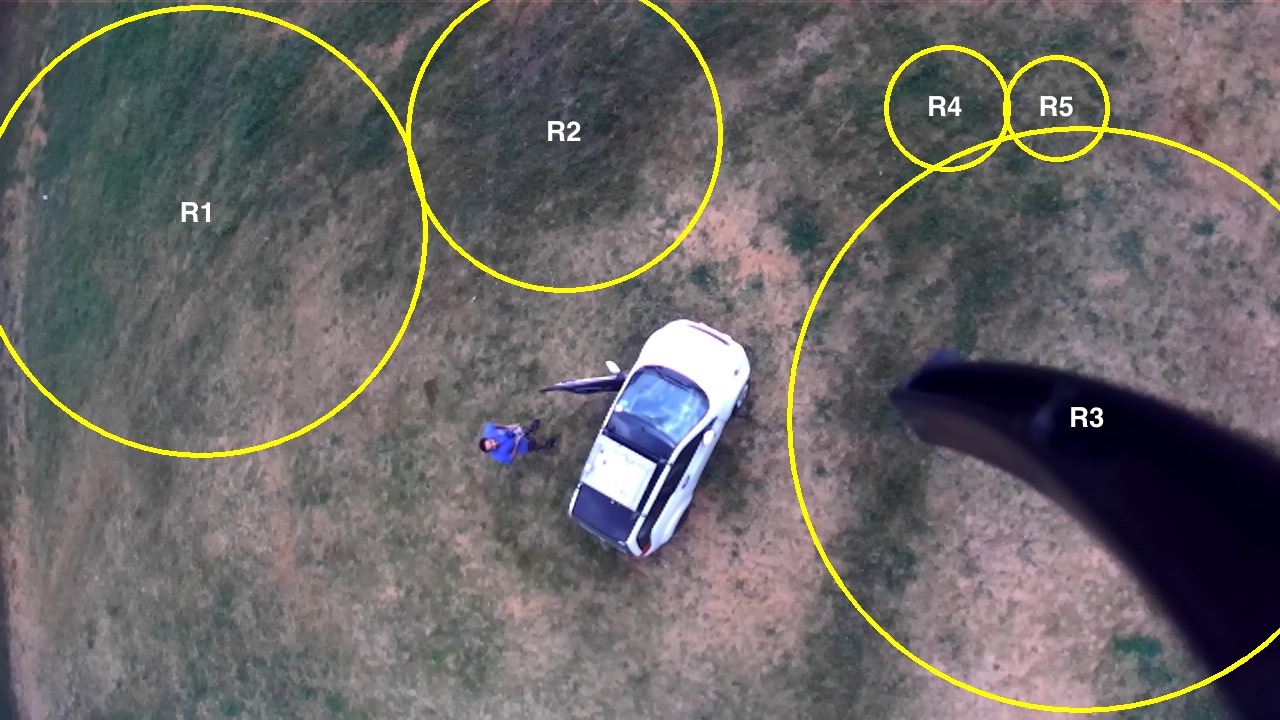}  
  \caption{Rural Scenario}
  \label{fig:sub-first1}
  \end{subfigure}
  \begin{subfigure}[b]{.4\textwidth}
  \includegraphics[width=\textwidth]{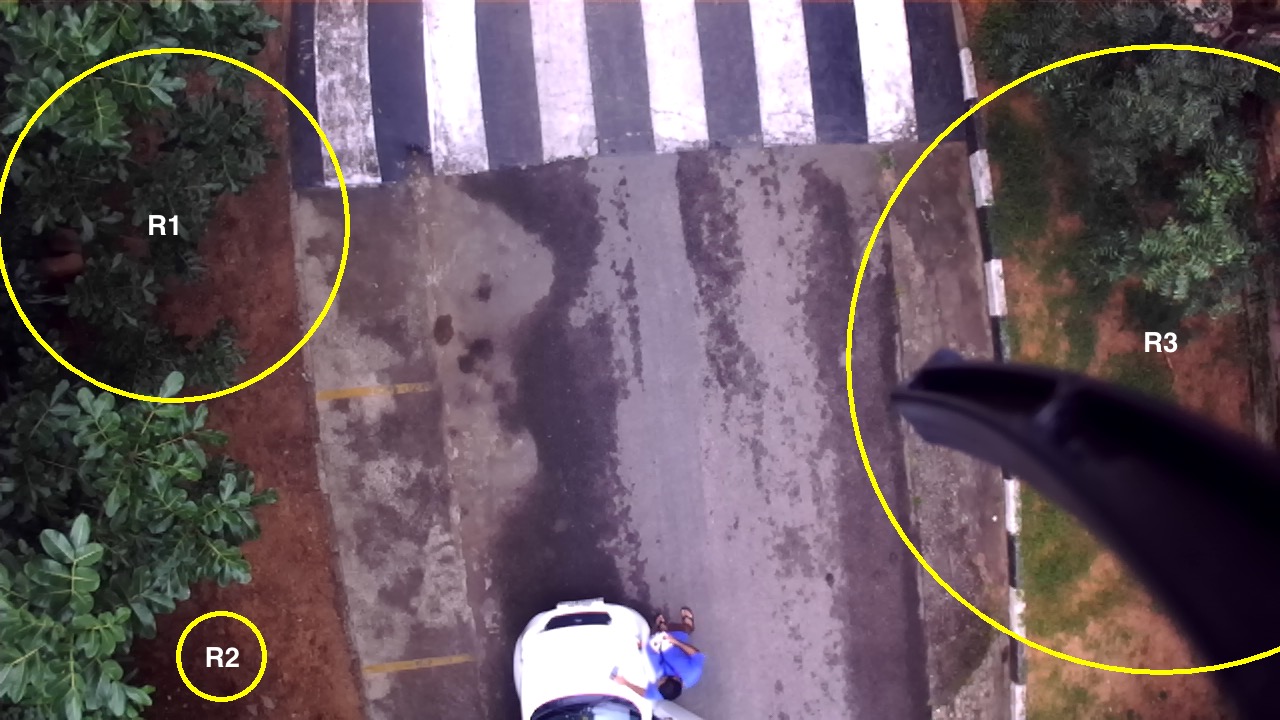}  
  \caption{Urban Scenario}
  \label{fig:sub-second2}
  \end{subfigure}
  
  \begin{subfigure}[b]{.4\textwidth}
  \includegraphics[width=\textwidth]{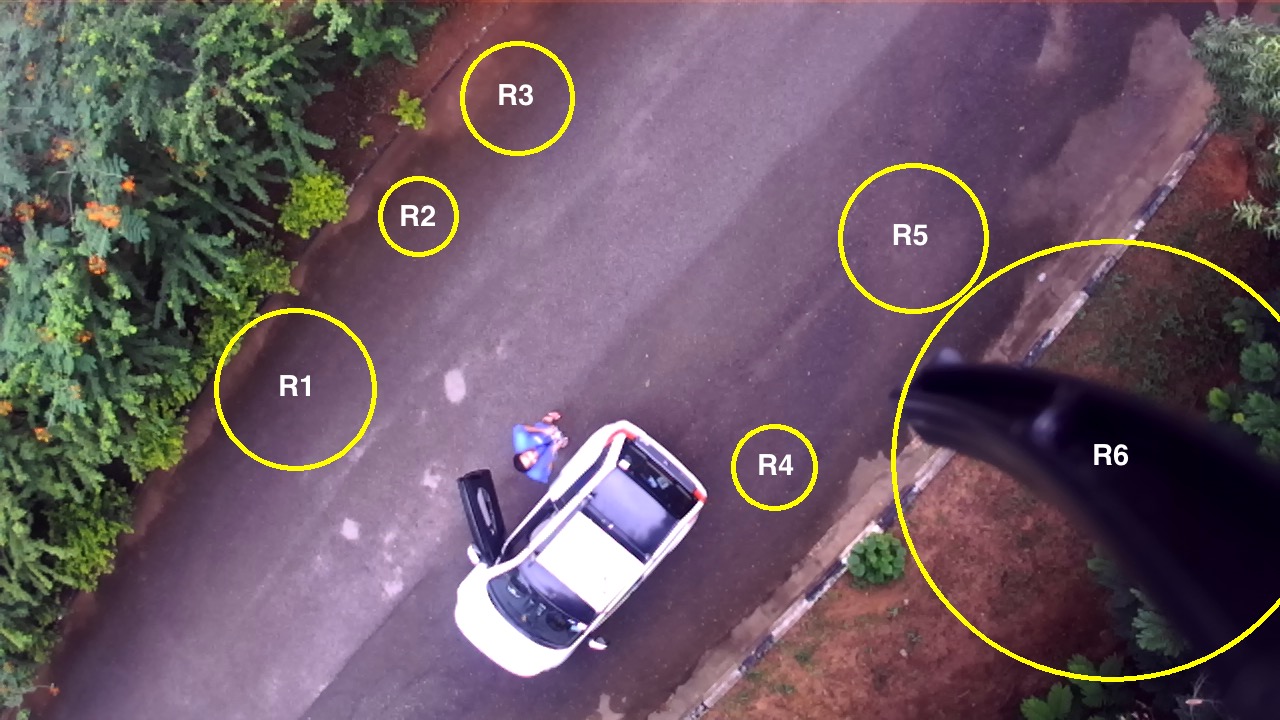}  
  \caption{Sub-Urban Scenario }
  \label{fig:sub-third3}
  \end{subfigure}

\caption{\ref{fig:sub-first1}, \ref{fig:sub-second2}, and \ref{fig:sub-third3}, represents the real-time area estimation through flying drone for the PLZ in different scenarios.}
\label{real-data-dynamic}
\end{figure*}

We validated our PLZ detection module in real-world scenarios. In particularly, we tested it in Rural, Urban, and Sub-Urban scenarios as shown in the Figure \ref{fig:sub-first1}. The circles in yellow corresponds to the PLZ identified and their resulting diameters (distances) along with the area occupied can be visualized in Table  \ref{table:distanceresultsbuilding12}. After identifying the area, algorithm only considers the area having value greater than 3(sq. m). We use a Time-of-Flight (ToF) sensor for obtaining the ground truth that has a maximum range of 60 meters. The average percentage error for our distance and area-estimation method is recorded as 0.9692\% and 1.9427\% respectively. To the best of our knowledge, we didn't find any work that calculates the resulting area for PLZ with UAVs without some state-of-art deep learning networks. We also compare the accuracy with that from Google Earth, which has been listed as $\leq 1\%$ in \cite{2015-01-1435}. It should be noted that due to a lack of 3D imagery, some diameters that have been used for identifying the area cannot be measured using Google Earth. We also compare our area-estimation resultant average percentage error with the work done by authors that claims for $\leq 5\%$ error in \cite{9086493}.

  
  


\setlength{\tabcolsep}{1.5pt}
\newcolumntype{M}[1]{>{\centering\arraybackslash}m{#1}}

\begin{table*}

\caption{Landing area calculated for different PLZ using Our Method and Ground Truth for Rural, Urban, and Sub-Urban Scenarios through Drone.}

\label{table:distanceresultsbuilding12}
\centering
\begin{tabular}{|M{1.8cm}|M{2.2cm}|M{1.9cm}|M{1.9cm}|M{1.9cm}|M{1.9cm}|M{1.9cm}|M{1.9cm}|}
\hline
Scenarios &Reference &Estimated Distance (m)&Ground Truth Distance (m) &Distance Error (\%) &Estimated Area (sq. m)&Ground Truth Area (sq. m)&Area Error (\%) \\\cline{1-8}
\multirow{5}{*}{Rural} &R1 in Fig. 5(a) &3.8976 &3.9165 &0.4826 &11.9312 &12.0472 &0.9629 \\\cline{2-8}
&R2 in Fig. 5(a) &2.7144 &2.71 &0.1624 &5.7868 &5.768 &0.3259 \\\cline{2-8}
&R3 in Fig. 5(a) &5.0808 &5.125 &0.8624 &20.2747 &20.629 &1.7175 \\\cline{2-8}
&R4 in Fig. 5(a) &1.0674 &1.045 &2.1435 &0.8948 &0.8577 &4.3255 \\\cline{2-8}
&R5 in Fig. 5(a) &0.9046 &0.907 &0.2646 &0.6427 &0.6461 &0.5262 \\\cline{1-8}
\multirow{3}{*}{Urban} &R1 in Fig. 5(b) &3.065 &3.078 &0.4224 &7.3782 &7.4409 &0.8426 \\\cline{2-8}
&R2 in Fig. 5(b) &0.7492 &0.743 &0.8345 &0.4408 &0.4336 &1.6605 \\\cline{2-8}
&R3 in Fig. 5(b) &5.33 &5.286 &0.8324 &22.3123 &21.9454 &1.6719 \\\hline
\multirow{6}{*}{Sub-Urban} &R1 in Fig. 5(c) &0.7618 &0.753 &1.1687 &0.4558 &0.4453 &2.358 \\\cline{2-8}
&R2 in Fig. 5(c) &0.3714 &0.3745 &0.8278 &0.1083 &0.1102 &1.7241 \\\cline{2-8}
&R3 in Fig. 5(c) &0.532 &0.522 &1.9157 &0.2223 &0.214 &3.8785 \\\cline{2-8}
&R4 in Fig. 5(c) &0.4 &0.4112 &2.7237 &0.1257 &0.1328 &5.3464 \\\cline{2-8}
&R5 in Fig. 5(c) &0.6952 &0.7001 &0.6999 &0.3796 &0.385 &1.4026 \\\cline{2-8}
&R6 in Fig. 5(c) &2.0952 &2.1 &0.2286 &3.4478 &3.4636 &0.4562 \\\cline{2-8}
\hline
\end{tabular}
\end{table*}

\subsection{Real-Time Potential Landing Zone State Estimation and Navigation in Dynamic Scenarios}

We studied different scenarios as shown in Figure \ref{real-data-dynamic}, identifying different moving objects in the camera's field of view. We validated our algorithm, where we identified the speeds of the moving objects (vehicles) as shown in the Table \ref{table:velocity}. The ground truth was measured through the odometry in the vehicles. Hence, the results were averaged out with the percentage error of 2.37\%. To the best of our knowledge, we haven't found any work stating the speeds with the help of UAVs in dynamic scenarios. 

\begin{figure*}
  \centering
  
  \begin{subfigure}[b]{.4\textwidth}
  \includegraphics[width=\textwidth]{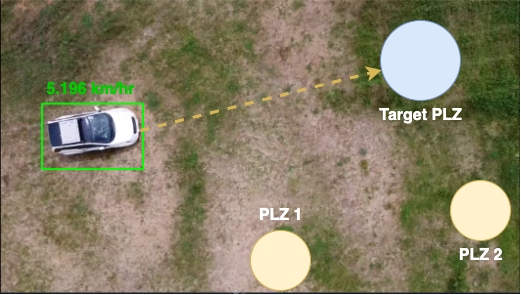}  
  \caption{Rural Scenario}
  \label{fig:sub-first}
  \end{subfigure}
  \begin{subfigure}[b]{.4\textwidth}
  \includegraphics[width=\textwidth]{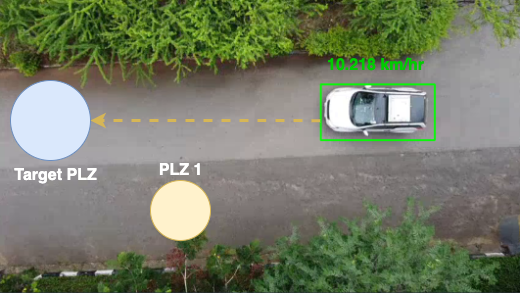}  
  \caption{Urban Scenario}
  \label{fig:sub-second}
  \end{subfigure}
  
  \begin{subfigure}[b]{.4\textwidth}
  \includegraphics[width=\textwidth]{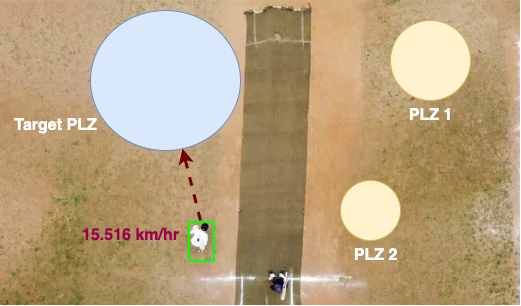}  
  \caption{Sub-Urban Scenario }
  \label{fig:sub-third}
  \end{subfigure}

\caption{\ref{fig:sub-first}, \ref{fig:sub-second}, and \ref{fig:sub-third}, represents the real-time velocity estimation through flying drone for the PLZ in different scenarios.}
\label{real-data-dynamic}
\end{figure*}

\setlength{\tabcolsep}{1.5pt}
\newcolumntype{M}[1]{>{\centering\arraybackslash}m{#1}}
\begin{table*}
\caption{Velocity Estimation through translating the pixels. Ground truth has been measured through Accelerometer}
\label{table:velocity}
\centering
  \begin{tabular}{|M{1.8cm}|M{2.4cm}|M{1.7cm}|M{1.7cm}|M{1.7cm}|}

    \hline
  Scenarios & Velocity Measured using Accelerometer (Km/hr) & Estimated Velocity (Km/hr) & Absolute Difference & Percentage Error\\
  \hline
  Scene 1 & 5 & 5.196 & 0.196 & 3.92 \\
  \hline
  Scene 2 & 10.2 & 10.218 & 0.018 & 0.17 \\
  \hline
  Scene 3 & 16 & 15.516 & 0.484 & 3.025 \\
  \hline
  \end{tabular}
\end{table*}
\section{DISCUSSION}
This is an ongoing research through which we are trying to focus more on complex dynamic scenarios, taking into consideration the urban areas where the quadrotor has to navigate and land safely in the case of aborted missions under various applications. So far, the results presented in this paper have been tested and evaluated in real-world, also compared with some existing work. We released an open source framework which can be embedded in any microprocessor and could be used for future purposes. Additional details and code can be accessed at \url{https://github.com/jaskiratsingh2000/Research-UAVs-Safe-Landings}

\section{CONCLUSION}
In this study, a large number of strategies that have been devised to ensure the safe landing of UAVs in dynamic environments while taking into account moving objects were put into practise. Safe landing of UAVs involves estimation of different safe potential landing zones. Hence, it is important to estimate the parameters that could identify safe PLZ, and navigate to land there. In particular, we identify PLZ using canny-edge detection, which is further used to calculate the required area using nearest neighboring contours. Using diameter-area calculation, we did real-time state estimation and navigation in dynamic scenarios. We evaluated all our results on the real-world data, aiming at better approaches to land safely in dynamic scenarios.
Future work includes considering the estimating the trajectories of the moving objects, and considering the run-time evaluation while drone itself is in dynamic conditions.

\newpage

\bibliographystyle{IEEEtran}
\bibliography{root}

\end{document}